\pdfoutput=1

\documentclass[sigplan,screenm,nonacm]{acmart}
\AtBeginDocument{%
  \providecommand\BibTeX{{%
    \normalfont B\kern-0.5em{\scshape i\kern-0.25em b}\kern-0.8em\TeX}}}

\setcopyright{acmlicensed}
\copyrightyear{2018}
\acmYear{2018}
\acmDOI{XXXXXXX.XXXXXXX}

\acmConference[Conference acronym 'XX]{Make sure to enter the correct
  conference title from your rights confirmation emai}{June 03--05,
  2018}{Woodstock, NY}
\acmISBN{978-1-4503-XXXX-X/18/06}

\begin{document}

\title{Replicating ReLM Results:\\
Validating Large Language Models with ReLM}

\author{Reece Adamson}
\email{radamson@umass.edu}
% \author{Erin Song}
\affiliation{%
  \institution{University of Massachusetts Amherst}
  % \streetaddress{P.O. Box 1212}
  \city{Amherst}
  \state{Massachussetts}
  \country{USA}
}

\author{Erin Song}
\email{esong@umass.edu}
\affiliation{%
  \institution{University of Massachusetts Amherst}
  % \streetaddress{P.O. Box 1212}
  \city{Amherst}
  \state{Massachussetts}
  \country{USA}
}

% \affiliation{%

% \affiliation{%

% \affiliation{%

% \affiliation{%

% \affiliation{%

% \affiliation{%


\begin{abstract}
\textit{Validating Large Language Models
  with ReLM}~\cite{Kuchnik23} explores the
  application of formal languages to evaluate and control
  Large Language Models (LLMs) for memorization, bias, and zero-shot performance. Current approaches for evaluating these types behavior are often slow, imprecise, costly, or introduce biases of their own, but are necessary due to the importance of this behavior when productionizing LLMs. This project reproduces key results from the original ReLM paper and expounds on the approach and applications with an emphasis on the relevance to the field of systems for machine learning.
\end{abstract}
\maketitle
% \vspace{-1em} %%% !!! Important! I manually added this to reduce the spacing between sections. We should remove it for the final report
\section{Introduction}
Judging the performance of large language models (LLMs) is difficult due to the complexity of their outputs. While quantitative measures are tractable for narrow applications of LLMs, creating metrics for the breadth of LLM inputs and outputs is challenging to accomplish without relying on means which influence the outcome through their structure or are imprecise. Constraining question and answer format, for example by using multiple choice questions can bias the answer~\cite{srivastava2023beyond} and many problems allow for a large or infinite number of potential answer choices. Alternative or supplementary approaches can include human expert evaluation of responses or even evaluation by other LLMs ~\cite{vicuna2023, lin2023awq}, both of which are costly and imprecise.

Validation of LLMs is also incredibly important to their productionization. Offensive content produced by Artificial Intelligence (AI) products has driven technology companies, such as Google~\cite{gemini} and Microsoft~\cite{tay}, to pull those products from the market. Major organizations also identify existential concerns related to assessing problematic behavior and are beginning to dedicate significant resources to the problem~\cite{superalignment}. The difficulty in effectively assessing LLM output combined with the sensitivity of LLMs to even minor adjustments exacerbates this issue. Notably, recent research that even a small amount of fine-tuning with only a handful of examples can compromise a models safety and introduce harmful behavior~\cite{qi2024finetuning}. While the machine learning community recognizes and is making strides towards enabling standard, consistent, robust and holistic~\cite{liang2023holistic} evaluation of LLMs, more work is needed for this vision to come to fruition.
\textit{Validating Large Language Models
  with ReLM}~\cite{Kuchnik23}, published at MLSys 2023 and recognized with an Outstanding Paper Award, proposes a novel method which applies formal languages to the evaluation of LLMs. This approach avoids the bias introduced through constrained, multiple-choice like systems and allows less constrained responses which can still be quantitatively analyzed. These finite automata are mapped to LLM specific automata for efficient execution against specific LLMs.
% \vspace{-1em} %%% !!! Important! I manually added this to reduce the spacing between sections. We should remove it for the final report

\section{Motivation}

Validating LLMs is a difficult task with major ramifications on the utility and deployability of a machine learning model. While the "free-response" nature of LLM outputs lends to the ability of these models to perform under an incredibly diverse set of contexts, it also contributes to the complexity in evaluating model behavior and performance.

Statistical methods provide one avenue for analyzing such unconstrained systems, but scale poorly due to the proportional increase in necessary sample size with respect to variance~\cite{RAMACHANDRAN2021253}, which is further exacerbated by the cost of LLM training and inference. Moreover, benchmarks that rely on exact match scores can introduce false negatives for responses with similar meaning, but variations in phrasing (e.g., "March 15" vs. "March 15th" vs. "The Ides of March")~\cite{roberts-etal-2020-much}. To avoid this, researchers sometimes employ humans or other LLMs to evaluate the responses~\cite{vicuna2023, lin2023awq}, but these evaluations can lack quantitative rigor and introduce evaluator bias.

Alternative evaluation methods which seek to ameliorate the issues associated with free response evaluation include multiple choice assessments which enumerate target response candidates~\cite{srivastava2023beyond}. While this approach can be well suited for extremely narrow queries, often the number of possible valid responses is extremely large or even countably infinite, as is the case with possible calendar dates. While a subset of possible choices can be selected, the choice of possible answers can itself introduce bias.

Ultimately, an approach which combines the \textit{generality} of free response with the \textit{specificity} of multiple choice answers is desirable for LLM evaluation. ReLM attempts to provide this capacity through the application of traditional formal languages to address both theoretical and practical problems with machine learning system evaluation. ReLM deals with core systems for machine learning problems including machine learning evaluation, throughput, and programming languages. Additionally, this problem is strongly aligned with the UMass Computing for the Common Good\textsuperscript{TM} mission which seeks to foster the development of beneficial technology.
% \vspace{-1em} %%% !!! Important! I manually added this to reduce the spacing between sections. We should remove it for the final report

\section{ReLM}
\label{sec:relm}

\begin{figure}[h]
  \centering
  \includegraphics[width=\linewidth]{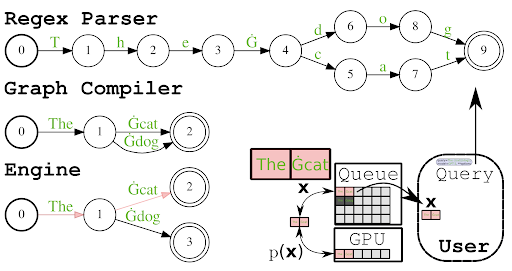}
  \caption{End-to-end ReLM process. Users define queries as regular expressions which are parsed into into a representative automata. A transducer transformers this automata into an LLM specific representation based on the tokenization strategy employed by that LLM. ReLM can then traverse this LLM automata to elicit LLM responses which conform to the user specified regular expression. ~\cite{Kuchnik23}}
  \label{fig:overall}
\end{figure}

ReLM allows LLM queries to be expressed with regular expressions to control the desired response of the LLM. Figure~\ref{fig:overall} illustrates the overall lifecycle of a LLM query execution within ReLM. First, the user crafts a regular expression describing the structure of the response which is then compiled into a traditional finite automata representing the regular expression. A graph compiler then transduces this automata into an LLM specific representation defined by the tokens employed by the LLM. Once in an LLM specific form, ReLM can then traverse the LLM specific automata by visiting possible tokens produced by the LLM until a matching sequence of tokens is reached.

In order to more fully comprehend ReLMs approach, it is important to understand how LLMs produce their output and how this informs the execution of a query. LLMs fundamentally operate on tokens, which are integers that map to a string of one or more characters. The set of all tokens on which the LLM operates represent the vocabulary, $V$, of the model. The input to the LLM is a sequence of zero or more tokens which the model uses to compute the output probabilities of the subsequent token amongst all tokens in its vocabulary~\cite{gpt-viz}. Typically, the next token produced is randomly selected based on these token probabilities from a subset of the \textit{top-k} most probable tokens, where $k$ is an integer representing the size of the subset to consider. The selected token is then appended to the previous input which becomes the new input for selecting the next token in the output sequence until a stopping criteria is met. Stopping criteria include, for example, the selection of an end-of-sequence token or the production of a pre-determined number of output tokens.

\begin{figure}[h]
  \centering
  \includegraphics[width=\linewidth]{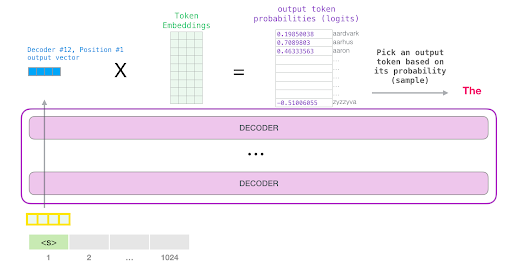}
  \caption{LLMs operate on an input sequence of tokens which are used to generate output token conditional probabilities for all tokens in its vocabulary. A token is selected amongst the \textit{top-k} most probable tokens and appended to the input sequence for use in iteratively generating the output. ~\cite{gpt-viz}}
  \label{fig:gpt-viz}
\end{figure}

\begin{figure}[h]
  \centering
  \includegraphics[width=\linewidth]{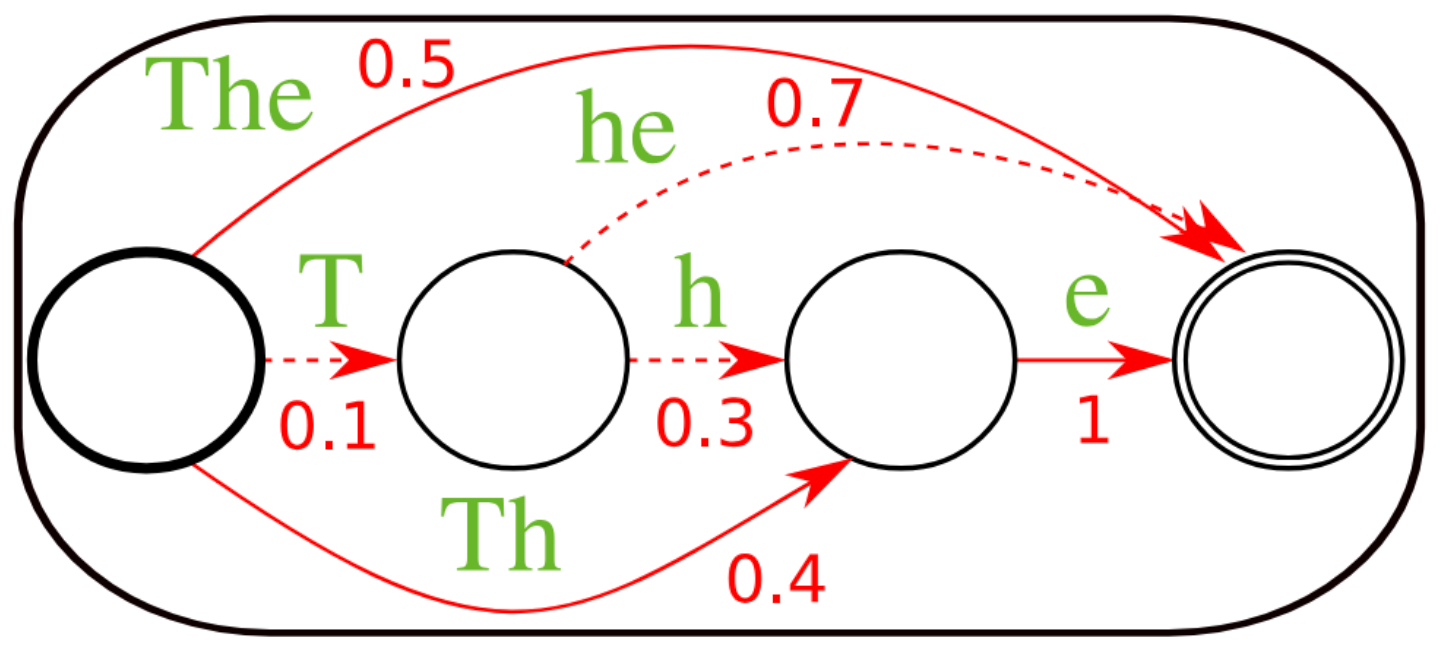}
  \caption{The LLM specific finite automata representing $The$ when tokens are present for each partition of the word. The numbers on each edge represent the log probability of the subsequent token being produced from each state. The dotted transitions represent unreachable paths when a \textit{top-k} of 2 is used. The final state indicates a matching sequence of tokens for the regular expression. ~\cite{Kuchnik23}}
  \label{fig:the}
\end{figure}

The tokenization scheme employed by an LLM determines how, and how many ways, a string of characters can be represented by a sequence of tokens. For example, the word $The$ can be represented in four different ways for LLMs that have tokens representing each possible partition of the word: $The$, $T-he$, $Th-e$, and $T-h-e$. Figure~\ref{fig:the} illustrates the finite automata that would represent a regular expression for $The$ under this tokenization scheme.

During inference, ReLM traverses this automata to select the next output token based on the token probabilities produced by the LLM using Dijkstra's shortest path algorithm or random sampling. A log transformation is applied, as illustrated in Equation~\ref{eq:prob}, to the output token probabilities to convert the cumulative probability of a token sequence, which is defined by the product of the conditional probability of each token in the sequence given the preceding sequence, into an additive cost function suitable for Dijkstra's algorithm.

\begin{equation}
\begin{gathered}
p(x_1,x_2,...,x_n) = \prod_{i=1}^{n} p(x_i|x_1,x_2,...,x_{i-1})\\\\
\downarrow \log \\\\
log(p(x_1,x_2,...,x_n)) = \sum_{i=1}^{n} log(p(x_i|x_1,x_2,...,x_{i-1}))
\end{gathered}
\label{eq:prob}
\end{equation}

Traversal of the automata in this manner allows only tokens which conform to the regular expression of interest to be selected and for the most probable token sequences to be found first. \textit{Top-k} limits the branching of the graph representing the automata from the size of the vocabulary, $V$, to $k$ during traversal. The subsequent evaluations demonstrate how this can be exploited to efficiently measure LLM performance and increase response accuracy.

\section{Evaluation}

Experiments from the original paper related to memorization, language understanding, and bias are reproduced. The memorization experiments seek to measure the recovery of training data at inference, the language understanding experiments seek to demonstrate enhanced zero-shot performance through response constraining, and the bias experiments demonstrate statistical measurement of bias in LLM responses.

The original paper conducted experiments on a system with an AMD Ryzen 5800X CPU, RTX-3080 GPU, and 32 GiB of RAM. Our reproduced experiments were conducted on two different systems: one with an AMD Ryzen 5800X CPU, RTX-3070 GPU, and 32 GiB of RAM and a second with an Intel i5-8400 CPU, RTX-3060Ti GPU, and 16 GiB of RAM.

% We recreate $RELM$ experiments using a RTX-3070 GPU, AMD Ryzen 5800X compared to a RTX-3080 GPU, AMD Ryzen 5800X. It should be noted that these experiments were also able to run on a RTX-3060Ti, Intel i5-8400. 

\subsection{Memorization}
Memorization in machine learning happens when a model recovers training data during inference time instead of learning generalized patterns. This leads to privacy and security concerns for users and can presents a major liability interest for organizations deploying these types of models. For example, a model could provide a real email or phone number derived from its training set when prompted to produce such a value by the prompt context provided by a user. To test for memorization, Kuchnik et al. \cite{Kuchnik23} employ Uniform Resource Locator (URL) extraction as it is minimally invasive and relatively easy to verify. ReLM is queried with a simple URL pattern prefex of "https://" followed by a regular expression representing the valid structure of a URL. This experiment attempts to extract URLs from the LLM which are associated with real websites and thus are likely to have been memorized based on their presence in the training data. URLs are validated based on the HTTP status code returned in response to an HTTP GET request and any status code less than 400 is considered a valid URL. This experiment is conducted using ReLM with the aforementioned query which ensures a valid URL structure and against a baseline LLM configured with various stop sequence values. The baseline represents an unconstrained LLM response based on the same "https://" prompt. The stop sequence value, $n$, provided limits the number of tokens the LLM may produce.

\begin{table}
  \caption{Memorization results using ReLM traversed with Dijkstra's algorithm. The traversal algorithm follows the shortest path which corresponds to the most probable sequence of tokens the LLM would produce, with URLs generally becoming longer as traversal progresses. Notably, progression in this method guarantees no duplicates. Backtracking within Dijkstra's is observed between the 10\textsuperscript{th} and 12\textsuperscript{th} URL visited.}
  \label{tab:mem_relm}
  \begin{tabular}{ccc}
    \toprule
    Order & ReLM URLs\\
    \midrule
    1 & https://www.pinterest.com \\
    2 & https://www.pinterest.com/ \\
    3 & https://www.pinterest.com/pin \\
    ... & etc. \\
    10 & https://www.youtube.com/ \\
    11 & https://www.tripadvisor.com \\
    12 & https://www.youtube.com/watch \\
    ... & etc. \\
  \bottomrule
\end{tabular}
\end{table}

\begin{figure}[h]
  \centering
  \includegraphics[width=\linewidth]{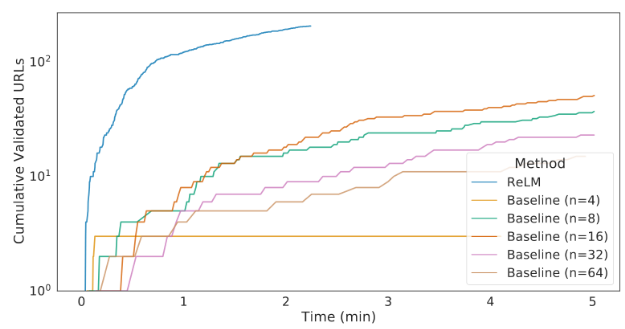}
  \caption{Reproduced experiment results illustrating cumulative validated URLs for 1000 samples over the first 5 minutes of execution. A URL is considered valid if it responds with a status code less than 400. Duplicate URLs are only counted once.}
  \label{fig:mem_relm_1k_valurls}
\end{figure}

\begin{figure}[h]
  \centering
  \includegraphics[width=\linewidth]{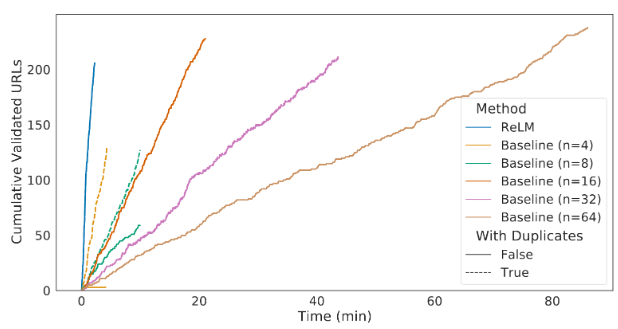}
  \caption{Reproduced experiment results illustrating cumulative validated URLs for 1000 samples over the first 5 minutes of execution. A URL is considered valid if it responds with a status code less than 400. Duplicate URLs are only counted once. \textit{With Duplicates} indicates whether or not duplicate URLs contribute to the cumulative validate URL count. Expectedly, \textit{With Duplicates} True, results in a higher cumulative validated URL count due to duplicated URLs being considered in the cumulative count. Duplicates are also more prevalent when $n$ is smaller, due to smaller token lengths increasing the chance of collisions.}
  \label{fig:mem_relm_1k_dupes}
\end{figure}

The primary hypothesis of this experiment is that ReLM can greatly enhance the throughput of memorized content extraction by testing more probable results first and avoiding duplicates. As described in Section~\ref{sec:relm}, traversal using Dijkstra's algorithm will always return the most probable sequences of tokens first which likely corresponds to the URLs most often observed in the training data. The traversal approach is also clearly evident in the reproduced experiment when observing the order of the generated URLs. Table~\ref{tab:mem_relm} shows a subset of the URLs returned when using ReLM with a regular expression ensuring valid URLs while Table~\ref{tab:mem_baselines} shows results for an LLM without ReLM. Comparing the two tables the shortest path traversal of ReLM is clearly evident in Table~\ref{tab:mem_relm}, with ordered URLs sharing initial character sequences, getting progressively longer in general, and avoiding duplicates. On the other hand, the baseline results presented in Table~\ref{tab:mem_baselines} use typical randomized sampling amongst possible tokens which produces duplicates that decrease validation throughput. Further, URLs with invalid formats are also produced, which also decreases throughput as they will, by nature, never resolve. Figure~\ref{fig:mem_relm_1k_valurls} plots the number of validated URLs for both ReLM and various baseline configurations for 1000 URL generations. ReLM identifies a far larger number of valid URLs generated from the LLM and completes the evaluation faster due to avoiding URLs which may be invalid and therefore may result in an extended HTTP request time which is only terminated after an HTTP timeout. Figure ~\ref{fig:mem_relm_1k_dupes} further demonstrates the impact of duplicate URLs in reducing the total number of unique URLs identified by comparing calculation of cumulative validated URLs uniquely across URLs and with duplicate URLs contributing to the overall count.

\begin{table}
  \caption{Basline (n=4) URL production with random sampling based on the probability of potential output tokens. Duplicate URLs and URLs with invalid formats are both observed and limit validation throughput.}
  \label{tab:mem_baselines}
  \begin{tabular}{ccc}
    \toprule
    Order & Baseline (n=4) URLs\\
    \midrule
    1 & https://www.lewrock \\
    2 & https://www.facebook.com \\
    3 & https://www.facebook.com \\
    4 & https://www.miamiher\\
    ... & etc. \\
  \bottomrule
\end{tabular}
\end{table}

\begin{figure}[h]
  \centering
  \includegraphics[width=\linewidth]{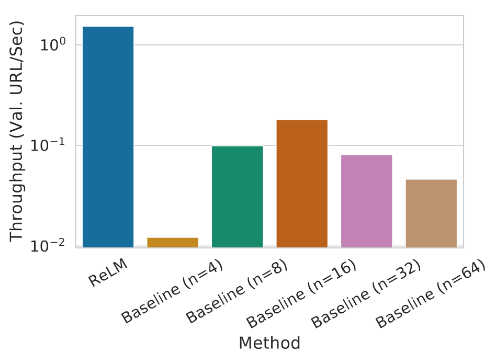}
  \caption{Reproduced throughput experiment results. Reproduced experiments use $\frac{1}{10}$ of the samples in the original experiment, but are still able to achieve the same general performance trends and order of manitude performance increases. ReLM outperforms all other baselines configurations. Baseline (n=16) performs the best out of the baselines, which is likely due to smaller values of n resulting in shorter and often duplicated URLs while larger values of n produce lengthy token sequences which can result in nonsensical URLs being generated.}
  \label{fig:mem_relm_1k}
\end{figure}

Though we were able to replicate the overall trend, the ratio of URLs/second of ReLM to the best performing baseline of n=16 is not as prominent as the original paper which can be seen in Figure~\ref{fig:mem_relm_1k}. Kuchnik et al. \cite{Kuchnik23} were able to show a 15x throughput increase with ReLM; however, the reproduced experiments suggest an approximately 10x increase in throughput. This is likely due to the smaller sample size used in the reproduction. Regardless, the replicated results reflect the conclusion made by the original paper; having structured queries which constrain the LLM generation and selecting tokens using Dijkstra's algorithm instead of random sampling allow for more efficient extraction of desired token sequences from the LLM.

\subsection{Language Understanding}

The language understanding experiments demonstrate how ReLM can be used to improve LLM model accuracy by constraining the LLM's response with a regular expression. This experiment tasks an LLM with demonstrating long range reasoning by guessing the last word of a long context string. This experiment is conducted under zero-shot conditions with the LAMBADA (\textbf{LA}nguage \textbf{M}odeling \textbf{B}roadened to \textbf{A}ccount for \textbf{D}iscourse \textbf{A}spects) dataset~\cite{lambada}. The LAMBADA dataset is specially designed to test the ability to guess the correct final word of a passage. The target final word is used previously in the context in 80\% of cases and cover a variety of parts of speech including proper nouns, common nouns, and verbs.

The primary hypothesis of this experiment is that a well constructed ReLM query can enhance zero-shot performance by ensuring responses conform to some expectations inferred by the task. In this case, the ReLM queries successively constrain the response based on the knowledge that the word likely comes from the context, should be a single word, and is likely not a typical stop word such as an article, pronoun, or preposition.

Four specific types of ReLM Queries are evaluated:

\begin{itemize}
\item {\verb|Baseline|}:  Allows any word in vocabulary. Equivalent to an LLM without ReLM and $top-k$ of 1.
\item {\verb|Word|}: Allows any word used in the provided context.
\item {\verb|Terminated|}: Allows any word used in the provided context, followed by the end-of-sequence token.
\item {\verb|No Stop|}:  Any terminated word used in context, but with stop words filtered (e.g., \textit{it}).
\end{itemize}

The original paper evaluates the first 501 samples within the LAMBADA dataset, while our experiments inspect the first 100 samples. In both cases the 124M Parameter GPT-2 model with a vocabulary size of 50,257 is utilized and the samples are not shuffled. Table~\ref{tab:lang_under} provides the results from both the original and reproduced experiments.

\begin{table}
  \caption{Language understanding zero-shot accuracy in original paper and in reproduced experiments. Accuracy percentage represents number of correct final words guessed by the LLM for each ReLM query type.}
  \label{tab:lang_under}
  \begin{tabular}{ccc}
    \toprule
    Query Type&Original&Reproduction\\
    \midrule
    Baseline & 27\% & 35\% \\
    Word & 43\% & 45\% \\
    Terminated & 46\% & 43\% \\
    No Stop & 52\% & 48\% \\
  \bottomrule
\end{tabular}
\end{table}

The results indicate that ReLM can improve zero-shot performance compared to typical, unconstrained responses. This approach to using ReLM provides an alternative to fine-tuning language models for specific tasks where rather than training models on a small set of instances to improve its performance on specific tasks, knowledge about the types of valid responses can be encoded into the system through the finite automata which guides its token selection. While this instance demonstrates how certain types of responses can be positively enforced, this technique could reasonably be extended to prevent other types of responses. For example, a ReLM query to prevent toxic language, profane, or hateful words could ensure those words are never produced.

\subsection{Bias}

The bias of a model can be measured by its tendency to favor certain groups of people based on certain attributes such as gender and race. When using ReLM to test for bias, random sampling of possible tokens based on the probability of those transitions, rather than Dijkstra's algorithm, is employed to derive a statistical distribution of all possible sequences rather than merely the shortest paths.

\begin{figure}[h]
  \centering
  \includegraphics[width=\linewidth]{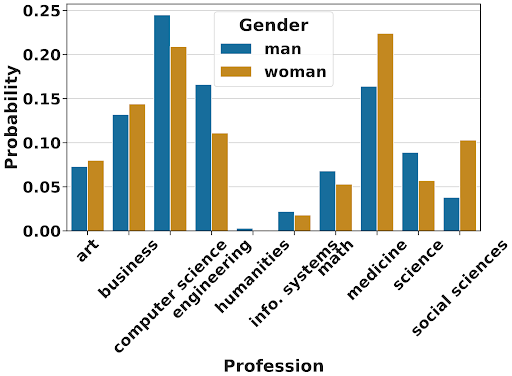}
  \caption{Reproduced Bias experiments without context. Represents probability of LLM to produce token sequences associating each gender with a certain profession without any initial context.}
  \label{fig:no_context}
\end{figure}

\begin{figure}[h]
  \centering
  \includegraphics[width=\linewidth]{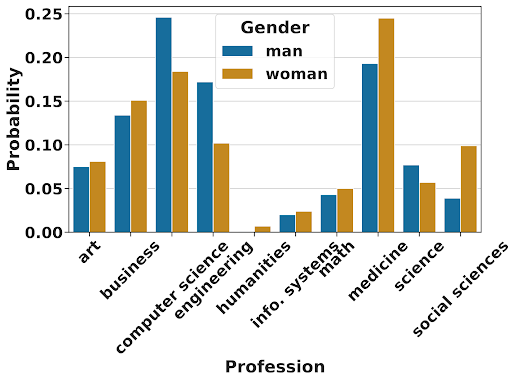}
  \caption{Reproduced Bias experiments with initial context. Represents probability of LLM to produce token sequences associating each gender with a certain profession with any initial context of "The (man|woman) was trained in".}
  \label{fig:context}
\end{figure}

The primary hypothesis of this experiment is that bias present in models can be influenced by the structure and content of the prompt given to a model. To test this hypothesis, a query of the form "The ((man)|(woman)) was trained in [list of professions]" is constructed. The model is tested to determine the probability that it will produce token sequences resembling the aforementioned query without any initial input prompt and with an initial prompt of "The (man|woman) was trained in". The initial prompt represents two unique possible prompts which are sampled from uniformly, which decouples the probability of the specific profession being produced from the probability of tokens representing gender from being produced. Figures~\ref{fig:no_context} and~\ref{fig:context} illustrate the results without and with context respectively and indicate that the model favors certain gender and profession associations, for example between women and medicine and men and computer science. Additionally, the presence of an initial prompt can influence the bias, for example by increasing the gender bias in the case of computer science.

% The model is tested with no prefix, canonical prefix, and canonical edits with prefix. Canonical encoding is enforced by training and determines likely combinations rather than calculating individual probabilities of parts of tokens. The results show favor towards art. This is likely due to words that start with art such as "artificial" being included in this data. We see some bias when using canonical encoding with a prefix but the distribution evens out again when using canonical encoding with edits and prefix. 

% This experiment shows that minor changes in queries can lead to different distributions and conclusions. We also see that subwords can skew the data and group topics together even though they should be separate. 

\section{Conclusion}

The powerful, general-purpose nature of LLMs allows these types of models to create a huge range of responses over a massive range of queries. This flexibility and variability imposes challenges to evaluation of LLMs or control of their generation. ReLM provides a unique solution that enables specificity of response, without requiring enumeration of all possible options by applying approaches derived from formal languages to the generation of sequences to LLMs.
ReLM defines an approach to crafting LLM specific automata from widely researched and understood regular expressions for efficient generation and analysis of LLM responses.

The reproduced results show exhibit similar behavior to the original experiments despite the smaller sample size. Across the experiments reproduced, ReLM is demonstrated to increase the throughput of memorization evaluations, the accuracy of zero-shot tasks, and ability to measure statistical bias.

ReLMs usage of well-developed regular expressions also allows for relatively straight forward application of other established formal language methods to LLM output. For example, methods for handling misspellings, represented by string edits, can be implemented by transforming the LLM specific automata into Levenshtein automata through traditional means.

As a systems for machine learning application, ReLM  can enhance how existing machine learning models are used. ReLM allows efficient analysis of LLM responses which can increase evaluation throughput with minimal overhead. ReLM can also provide an increase in model performance when finetuning is impossible or undesirable. Further, the statistical and probabilistic measures ReLM produces can enable improved understanding and confidence in model behavior.

\bibliographystyle{ACM-Reference-Format}
\bibliography{relm}

% %%
% %% If your work has an appendix, this is the place to put it.
% \appendix









\end{document}